\documentclass[letterpaper]{article} 
\usepackage{aaai24}  
\usepackage{times}  
\usepackage{helvet}  
\usepackage{courier}  
\usepackage[hyphens]{url}  
\usepackage{graphicx} 
\usepackage{natbib}  
\usepackage{caption} 
\usepackage{algorithm}
\usepackage{algorithmic}

\usepackage{newfloat}
\usepackage{listings}
\DeclareCaptionStyle{ruled}{labelfont=normalfont,labelsep=colon,strut=off} 
\floatstyle{ruled}
\newfloat{listing}{tb}{lst}{}
\floatname{listing}{Listing}
\usepackage[nolist]{acronym}

\acrodef{FI}[FI]{feature importance}
\acrodef{GDPR}[GDPR]{General Data Protection Regulation}
\acrodef{EU}[EU]{European Union}
\acrodef{ECJ}[ECJ]{European Court of Justice}
\acrodef{XAI}[XAI]{eXplainable Artificial Intelligence}
\acrodef{LIME}[LIME]{Linear Interpretable Model-Agnostic Explanations}
\acrodef{SHAP}[SHAP]{SHapley Additive exPlanations}
\acrodef{LRP}[LRP]{Layer-wise Relevance Propagation}
\acrodef{gpsd}[GPSD]{General Product Safety Directive}
\acrodef{gpsr}[GPSR]{General Product Safety Regulation}
\acrodef{prodsg}[ProdSG]{German Product Safety Act}
\acrodef{prodhaftg}[ProdHaftG]{Product Liability Act}
\acrodef{ML}[ML]{Machine Learning}

\title{How Should AI Decisions Be Explained? \\ Requirements for Explanations from the Perspective of European Law}
\author {
    Benjamin Fresz \textsuperscript{\rm 1,2},
    Elena Dubovitskaya\textsuperscript{\rm 3},
    Danilo Brajovic\textsuperscript{\rm 1,2},\\
    Marco F. Huber\textsuperscript{\rm 1,2},
    Christian Horz\textsuperscript{\rm 3}
}
\affiliations {
    \textsuperscript{\rm 1}Fraunhofer Institute for Manufacturing Engineering and Automation IPA, Stuttgart, Germany\\
    \textsuperscript{\rm 2}Institute of Industrial Manufacturing and Management IFF, University of Stuttgart, Germany\\
    \textsuperscript{\rm 3}University of Giessen, Germany\\
    \{benjamin.fresz, danilo.brajovic\}@ipa.fraunhofer.de, \\
    \{elena.dubovitskaya, christian.horz\}@recht.uni-giessen.de,
    marco.huber@ieee.org
}

\begin{document}

\maketitle

\begin{abstract}
This paper investigates the relationship between law and eXplainable Artificial Intelligence (XAI).
While there is much discussion about the AI Act, which was adopted by the European Parliament in March 2024, other areas of law seem underexplored.
This paper focuses on European (and in part German) law, although with international concepts and regulations such as fiduciary duties, the General Data Protection Regulation (GDPR), and product safety and liability.
Based on XAI-taxonomies, requirements for XAI methods are derived from each of the legal fields, resulting in the conclusion that each legal field requires different XAI properties and that the current state of the art does not fulfill these to full satisfaction, especially regarding the correctness (sometimes called fidelity) and confidence estimates of XAI methods.
\end{abstract}

\section{Introduction}
The AI Act of the European Union (EU) mandates that the design of high-risk AI systems enables deployers to ``interpret a system's output" (Art. 13(1) and 13(3)b vii of the AI Act) and notes that, for example in the case of AI use in law enforcement the ``exercise of important procedural fundamental rights $[...]$ could be hampered, in particular, where such AI systems are not sufficiently transparent, explainable and documented`` (Recital 59 of the AI Act).
However, a challenge lies in the vagueness of these transparency requirements and what exactly constitutes “sufficiently" transparent and explainable.
While more details on this will hopefully be specified in standards such as the ISO/IEC CD TS 6254, other legal bases for the use of \ac{XAI} already exist. This paper reviews the legal requirements for explainability of AI systems based on European (and in part German) law, focusing on three legal domains and application areas. Therefore, our contributions can be summarized as follows:
\begin{itemize}
    \item We present an extended list of \ac{XAI} properties that can be used to formulate requirements.
    \item We describe legal bases for the use of \ac{XAI}---besides the well-known case of the GDPR, also fiduciary duties and product safety and liability---and derive requirements that \ac{XAI} methods need to fulfill in each of the considered areas of law.
    \item We point out open research problems and a possible way forward for the field of \ac{XAI} by embracing interdisciplinarity.
\end{itemize}
For this, we first provide a brief overview of common \ac{XAI} terminology in the next section and establish the choice of areas of law. After the description of related works in Section~\ref{sec:related-work}, the \ac{XAI} properties used in this paper are described in Section~\ref{sec:xai-properties}, with the distinction of method-specific properties and process-specific properties.
Based on the separation of decision-centric and model-centric approaches, Section~\ref{sec:decision-centric} and Section~\ref{sec:model-centric} describe the relevant legal fields and formulate a task for \ac{XAI} in each of the relevant application areas.
Based on this task, the requirements for \ac{XAI} methods are derived.
The paper closes with a discussion in Section~\ref{sec:discussion} and a summary in Section~\ref{sec:summary}. The paper structure and requirements for \ac{XAI} are also replicated in short in Table~\ref{tab:summary} in Section~\ref{sec:summary}. 

\subsection{Taxonomy}
We start by summarizing the taxonomy of explainability methods for AI briefly. For a full introduction we point the readers to \cite{Burkart2021, Molnar2022}.
At the highest level, explanations can be divided into local and global ones. Local explanations explain a particular decision (e.g., the loan was rejected due to a low income) while global explanations reveal how decisions are made in general (e.g., every applicant with an income above €100k is given a loan).
They are introduced in more detail below. It needs to be noted that most \ac{XAI} methods provide non-causal explanations, although special causal methods exist. 

\subsubsection{Local Explanations} \label{ssec:local-explanations}
Feature attributions are widely used to provide local explanations by quantifying the contribution of specific features to a decision.
For example, in the loan granting scenario, feature attribution could reveal that a short length of employment had the highest negative impact on the rejection, while an income of €75k was the strongest positive feature.
Methods like \ac{SHAP} \cite{Lundberg2017} and \ac{LIME} \cite{Ribeiro2016} are commonly used for computing feature attributions. Saliency maps, on the other hand, highlight important regions in images that contribute to specific decisions. They can be created using methods like \ac{SHAP} or \ac{LIME}, although usually methods specifically for neural networks---such as Grad-CAM \cite{Selvaraju2019}---are used. Counterfactual explanations are a type of local explanation that provide insight into decision-making without revealing the detailed functioning of the \ac{ML} model \cite{Wachter2017}. These explanations show a change in input features necessary to alter the outcome. For instance, a counterfactual explanation for the loan example could be “If you had worked at least five years, you would have been granted the loan.” Data attribution or sample-based explanations involve identifying influential samples from the training set that contributed to a decision. In the loan example, this could be interpreted as “Your loan was rejected because similar applicants from the training data defaulted on a loan,” potentially accompanied by such similar applicants.

\subsubsection{Global Explanations} \label{ssec:global-explanations}
Global explanations provide information about the workings of a model across its entire data range. The simplest form of global explainability is the usage of so-called white-box models such as decision trees. These models provide out-of-the-box insights into the decision making process and, hence, do not need any specific post-hoc explainability---given that they are small enough in size to be comprehensible to humans. Providing global explanations for black-box models is more difficult. Apart from partial dependence plots (PDPs), surrogate models provide a way of opening the black box to some degree. For this, either a black-box model is transformed into a white-box one \cite{Nguyen2020} or a white-box (surrogate) model is fitted on the output of the black-box model that shall be explained, thus showing how similar decisions could be made. As a further way to explain neural networks, especially in computer-vision tasks, concept-based explanations can be used. They usually provide an account of which concepts, e.g., certain patterns like stripes or dots, are especially relevant for the classification of single images (local concept-based explanations) or for the class or model under consideration in general (global concept-based explanations) \cite{BeenKim2018}. Other methods produce global feature attribution explanations \cite{Covert2020}, highlighting features with high predictive power. 

\subsection{Legal Necessity for the Use of XAI} \label{ssec:legal-necessity}
To date, most works about the link between law and \ac{XAI} focus on the GDPR or the AI Act. In reality, the relationship between law and \ac{XAI} is even closer. 
There are many situations in which the law requires explanations.
In this paper, we focus on the need for explanation in relations between private actors, although such a need also exists in relations between the citizen and the state (for example, in the adoption of automated administrative acts or in criminal justice).
In doing so, we will distinguish between the need for global and for local explanations in the sense of the taxonomy outlined above.
Whether global or local explanations are required depends on the rights, obligations and legally protected interests involved.
If, for example, a loan application is rejected based on the \ac{ML} calculation, the applicant will want to know the reasons for this in their specific case.
If a “right to explanation” really exists in this situation (e.g., under the GDPR), there is a legal need for (at least) a local explanation.
If, on the other hand, the safety of an \ac{ML} application is at stake, a global explanation may be desirable from a legal perspective.
More generally, it depends on whether the legal focus is on a single decision or the entire model.
For each of these two groups of cases, we will present representative subgroups or examples which, by their very nature, do not claim to be exhaustive.
The need for explainability of an individual decision is considered (1) from the perspective of the decision-makers, whereby we selected company directors as a representative group (company law) and (2) from the perspective of the decision recipients (data protection law).
When discussing the need for explainability of the entire model, the focus will be on (3) developers (product safety law) and (4) operators (product liability).
This choice can be mapped to the definition of stakeholders for whom explanations are created described by \citet{Tomsett2018InterpretableToWhom} as operators and executors (group 1), data-subjects and decision-subjects (group 2), and creators and examiners of AI systems (group 3 and 4).
Note that an often discussed use case for \ac{XAI} is criminal litigation, which is not explicitly described here, because the requirements there depend on the specific question asked.
Thus, for criminal litigation, all explanation types could be used, given the right litigation task. As an example, in one of the first cases of \ac{XAI}-use in litigation, feature importance explanations with varying degrees of access to model and data were used \cite{Fraser2022XAIlitigation}.

\section{Related Work} \label{sec:related-work}
As described before, most related works have concerned themselves with the link between law and \ac{XAI} regarding the AI Act and the GDPR.
In the following, some of these works and the understanding of transparency in law are summarized.

A broader discussion around the need for algorithmic transparency was started by the provisions of the GDPR, especially about whether a “right to explanation” exists \cite{Wachter2016} and how much information would be necessary to fulfill such a right.
\citet{Wachter2017} devised counterfactual explanations to give users actionable advise while keeping the exact workings of an \ac{ML} model a secret (see Section~\ref{ssec:local-explanations}).
\citet{Vorras2021} argue for a “right to explanation” with the implication of a “qualified transparency mechanism”, i.e., adapting explanations to the target audience and “considering the technical limitations of AI interpretability, the underlying legal and regulatory barriers, as well as the legitimate expectations of stakeholders”.
A similar sentiment can be found in \cite{Kuzniacki2023}, where Kuzniacki et al. argue for the construction of a global regulatory framework for the use of \ac{XAI} in tax law.
They additionally recommend that “tax lawyers and computer scientists should work together to help reach a consensus on legal and technical solutions that secure \ac{XAI}’s place in tax law”.
While our work is not based on tax law, it follows a similar idea as it aims to combine the information from legal experts and computer scientists to make progress towards a structured use of \ac{XAI} for legal purposes.
As of now, a gap between the legal and technical understanding can be discerned in many works, as echoed by \citet{Gyevnar2023}, who argue that this difference in understanding can also be seen in the EU AI Act.
They note that there seem to be multiple understandings of transparency and that the transparency mandated in the EU AI Act is different from purely technical transparency, which is the aim of most works in \ac{XAI}.
Also concerned with the different interpretations of transparency in the AI Act are \citet{Busuioc2022}. They stress that the explainability of AI systems, i.e., the presentation of specific---mediated---details of their workings, is different from the disclosure of the model itself, where only the second option allows for proper scrutiny of such systems.
Such explanations and thus, mediated information, could be used maliciously, as shown by \citet{Zhou2023}.
Summarizing, despite some works in that direction, the use of \ac{XAI} for legal purposes is still unclear. 

\section{Properties of XAI-Methods (Possibly) Relevant for their Legal Use}
\label{sec:xai-properties}
To be able to formulate requirements for \ac{XAI} systems, a taxonomy of possibly necessary properties of such systems needs to be defined.
 To date, multiple such taxonomies have been proposed, even prompting a “taxonomy of taxonomies” \cite{Speith.2022}.
 In this work, the taxonomy of \citet{Nauta2022} is used, as an explorative literature research rendered it the most complete and comprehensive. Additionally, the organization into 12 so-called “Co-Properties”, which were derived from scientific literature published until the end of 2020, lends itself to discuss legal requirements for \ac{XAI} methods and explanations, due to the properties being more specific than in other taxonomies. This set of properties is described below.
Additional to these \ac{XAI} method specific properties, further ones are formulated in Section~\ref{ssec:process-properties} for the process in which \ac{XAI} is used. A short overview of the used properties can be found in Table~\ref{tab:properties}.

\begin{table*}[bt]
\label{a-sec:properties}

\centering
\begin{tabular}{|p{3.5cm}|p{12.25cm}|}
    \hline
    Property & Definition \\
    \hline\hline
    Correctness & Describes how faithful the explanation is w.r.t. the black box.\\
    Completeness & Describes how much of the black box behavior is described in the explanation.\\
    Consistency & Describes how deterministic and implementation-invariant the explanation method is. \\
    Continuity & Describes how continuous and generalizable the explanation function is. \\
    Contrastivity & Describes how discriminative the explanation is w.r.t. other events or targets. \\
    Covariate Complexity & Describes how complex the (interactions of) features in the explanation are. \\
    Compactness & Describes the size of the explanation. \\
    Compositionality & Describes the format and organization of the explanation. \\
    Confidence & Describes the presence and accuracy of probability information in the explanation. \\
    Context & Describes how relevant the explanation is to the user and their needs. \\
    Coherence & Describes how accordant the explanation is with prior knowledge and beliefs. \\
    Controllability & Describes how interactive or controllable an explanation is for a user. \\\hline
    Consilience & Describes whether more than one \ac{XAI} method should be used. \\
    Computations & Describes when an explanation needs to be available, e.g. right when a prediction is shown or at a later point of time. \\
    Coverage & Describes what should be explained, a single prediction (local), an entire model (global) or the influence of training data samples (for predictions or the overall model behavior). \\
    Counterability & Describes whether a process needs to be implemented which allows end-users to object to AI decisions or explanations. \\
    Constancy & Describes whether explanations need to be presented in a format that can be saved for later examination.\\
    \hline
\end{tabular}
\caption{12 Co-Properties of \citet{Nauta2022} (table adopted from \cite{Nauta2022}), extended with additional process properties in the bottom part of the table.}
\label{tab:properties}
\end{table*}

\subsection{12 Co-Properties of Nauta} \label{ssec:nauta-properties}
\emph{Correctness} describes whether---or to what extent---an explanation conforms to the \ac{ML} model the explanation was created for. While this seems like a very basic feature of any explanation method, this property is not necessarily fulfilled, for example for sample-based methods like \ac{LIME} and \ac{SHAP}.
\emph{Completeness} addresses how much of the explanandum---either a model itself or a prediction---is captured within an explanation. The extreme cases would be to either present the entire \ac{ML} model with its mathematical operations or to just highlight one input feature as the main cause of a specific outcome. White-box models are by definition complete explanations of themselves.
\emph{Consistency} demands that identical inputs provide identical explanations, as some explanation methods are not deterministic in their behavior---especially sample-based methods. Inconsistent explanations might also result from the hyperparameter choices of specific methods \cite{Tritscher2023}.
\emph{Continuity} poses some similarity to Consistency, as it demands that small changes in the input or output of the model should also only result in small changes in the explanation. This can be illustrated for saliency maps for images: If an image is overlaid with noise, but the prediction of the model does not change, the explanation should also only change slightly.
\emph{Contrastivity} describes the discriminativeness of an explanation, as explanations are supposed to show why a decision was made in contrast to a different outcome that was not chosen, as argued by \citet{Miller2017Insights}.
\emph{Covariate Complexity} shows how complex features and their interactions in an explanation are. The simplest way to go about this would be to use the input features and ignore all feature interactions (and thus possibly lose out on Correctness), whereas for example simplified feature interactions such as linear or monotonous interactions could be one step more complex.
\emph{Compactness} denotes the overall size of an explanation, as too large explanations often overwhelm users, whereas too small explanations provide less information about a decision or the model behavior \cite{Hoffman2023}.
\emph{Compositionality} describes how an explanation is presented, especially its format, organization, and structure. Different ways of presenting explanations can improve the way the explanation is received, for example by providing higher-level information \cite{AlvarezMelis2018} or different terminology \cite{Swartout1993}.
\emph{Confidence} estimates can be used in two different ways to assist in the use of \ac{XAI}---either to show the confidence of the model in the original prediction as part of the explanation or in denoting the confidence that a given explanation is correct and thus, conforms to the decision/model at hand.
\emph{Context} of an explanation, especially of who the explainee is, could help to adapt an explanation to specific user needs, thus improving its practical use. One of the earlier mentions of context can be found in \cite{Miller2017BewareOfInmates}, which even described the development of \ac{XAI} methods only conforming to the needs of AI engineers and scientists as “inmates running the asylum.”
\emph{Coherence} addresses whether an explanation conforms to prior knowledge---to that of an explainee, general consensus, or other beliefs---thus making it easier to understand. While humans often use their prior knowledge to evaluate whether they deem an explanations as trustworthy or correct, Coherence is distinctly different from Correctness as it is described above, as an explanation can be correct (i.e., conform to the logic of a model) without conforming to prior beliefs. \emph{Controllability} describes how and to what extent an explanation can be interacted with. Also called interactivity, it seems as this can improve the understanding of an explanation \cite{Cheng2019} as well and also helps to address Context, as users could adapt the explanation to their needs as they see fit. 
It should be noted that there is currently no consensus on the best way to evaluate the properties discussed, as different evaluation methods (with many listed in \cite{Nauta2022}) may contradict each other \cite{Tomsett2019SanityChecks}.
Therefore, this paper will not attempt to formulate specific requirements for validating \ac{XAI} systems, such as through specific technical means or user studies.

\subsection{Process Properties} \label{ssec:process-properties}
The aforementioned properties are mainly formulated for \ac{XAI} methods themselves, while for legal purposes, requirements could also be formulated for properties of the implementation or of the process of creating explanations. Thus, five additional properties are proposed and discussed in the following. 

Since most current \ac{XAI} methods do not guarantee Correctness, using multiple methods can help, e.g., to spot errors in an application. Given their \emph{Consilience}, this can also be an indication of Correctness.
However, if explanations do not align, there is no established way to resolve such situations.
The \emph{Computations} in real-time systems should also be considered. In time-sensitive scenarios, explanations need to be provided from the start, while in other cases, longer runtimes (e.g., hours of runtime on a GPU-server) may be feasible.
The \emph{Coverage} also needs to be considered, potentially being single decisions (local), entire models (global) or the data. Local explanations could be helpful to base a decision on, whereas data attributions might be more helpful where the traceability of errors is relevant.
While, strictly speaking, not part of the process of creating and showing explanations to users, it could also be a legal requirement to implement a way for end-users to object to decisions made by a model or to object to explanations that are conceived as false by the concerned user (\emph{Counterability} of decisions/explanations).
A further requirement can be \emph{Constancy}, as it might be necessary to refer to a documented decision and its explanation to avoid liability. In addition to just documenting an \ac{XAI}-explanation, in cases of ambiguous explanations (e.g., when multiple saliency maps are used to try to find general rules about how a computer-vison model decides) it might be necessary to not just document the explanation itself but accompany it by the interpretation of the person who used it to make a decision.

\section{Legal Requirements: Decision-Centric} \label{sec:decision-centric}
As mentioned above, the legal need for local explainability can be defined (1) from the perspective of decision-makers or (2) from the perspective of those affected by the decisions.
The decision-makers, e.g., corporate directors and officers, are typically interested in avoiding liability for their own \ac{ML} application-based decisions. The persons whose data is processed by an \ac{ML} application and who are affected by the decision (e.g., loan or job applicants) will have an interest in questioning the recommendation/prediction of the \ac{ML} model, especially if it is disadvantageous to them.
In the following, the legal situation of the respective stakeholders is described.
From this description, tasks for \ac{XAI} can be formulated for each use case, which then---in combination with the legal bases---result in requirements regarding the used \ac{XAI} methods.

\subsection{Perspective of Decision Makers: Plausibility Check} \label{ssec:plausibility-check}
People increasingly relying on \ac{ML}-based decision support systems poses a challenge for various areas of law.
Company law, which we use as an example in this paper, deals with corporate fiduciaries (directors and officers of a company) who use \ac{ML} applications when making decisions.
From the legal point of view, the fiduciaries remain solely responsible for the decision made.
If they violated their fiduciary duties in making the decision and the company suffered damage as a result, they must compensate for the damage (fiduciary duty claims).
A fiduciary can only avoid this liability if they prove that they did not breach any fiduciary duties (duties of care and loyalty) when basing a decision on the \ac{ML} recommendation/prediction. 

Ways to achieve this are intensely discussed in legal literature. Many scholars consider even the use of black-box \ac{ML} applications by fiduciaries to be unproblematic if the monitoring is properly organized and the results are correct \cite{Petrin2019, JingchenZhao2021AIcorporate}, while others argue that the lack of explainability leads the fiduciary to make an unfounded decision, thereby breaching the fiduciary duty of care \cite{CowgerJr2023}.
In Germany, the fiduciary duty of care has been specified by the Federal Supreme Court (BGH) in the so-called ISION judgement \cite{BGH2011}.
The ISION rule applies when the fiduciary seeks external advice due to a lack of own expertise. It has three requirements: (1) the expert consulted must be an independent professional who is qualified for the issue to be clarified, (2) the fiduciary must provide the expert with a comprehensive presentation of the company's circumstances and disclose all necessary documents, (3) the fiduciary must subject the expert's advice to a careful plausibility check.
A breach of these requirements constitutes a breach of the corporate duty of care. 
Comparable regulation exists in various jurisdictions; for example, Section~141(e) of the Delaware General Corporation Law (Title 8, Chapter 1 of the Delaware Code) affords corporate directors a defense against liability only when the expert is chosen “with reasonable care”, and the director “reasonably believes” that the matter falls within the expert’s competence, relying in “good faith” on the expert's advice.

While a discussion of the second requirement---the possibility of comprehensive presentation of the company's circumstances and disclosure of all necessary documents to the AI system---is also interesting, it is deemed out of the scope of this paper.
In the context of \ac{XAI}, the last requirement---the plausibility check---is particularly important. There is a consensus in the legal discourse that the above requirements should be applied to the use of an \ac{ML} application \cite{Buchholz2023}.
However, it remains unclear how the plausibility checks are to
be carried out without the traceability of \ac{ML} decisions.
Most authors seem to recognize that, at least in the case of black-box models, sufficient traceability is not given and a plausibility check is therefore de facto impossible.
For this reason, attempts are made to replace the plausibility check with less stringent requirements, such as requiring the manager to familiarize themselves with the basic functions of \ac{ML} applications and to regularly monitor the operation of the applications \cite{Moslein2018}.
However, this does not enable effective and sufficient control of algorithmic decisions. Instead, decision-making authority shifts from human to machine.
This remains true even if one resorts to the standard repertoire of organizational obligations, such as the duty to document decisions, appoint a Chief Information Officer (CIO), establish internal company rules for dealing with AI (e.g., through a special compliance brochure), and provide training for managers and other staff. 

Some legal scholars draw the comparison to a human expert and argue that the plausibility check is not about explaining the internal mental processes that have led to the expert's opinion, but rather about comparing this opinion with market knowledge, experience, and intuition. Therefore, it is argued, explainability is not required for AI decisions \cite{langenbucher2023}.
This is not correct, because the plausibility check is not a pure comparison with other evidence, but rather the examination of the expert's report itself to determine whether the expert made incorrect assumptions, was missing any information, or whether the report contains contradictions or errors that can be recognized by a layperson.

In order to be able to use \ac{ML} models for fiduciary decisions, a ``true'' plausibility check of their decisions needs to be possible and could be facilitated through \ac{XAI}.
To enable such a plausibility check, \ac{XAI} systems need to provide certain properties, as discussed below. Note that properties, for which no relevant requirements are seen for this case, are left out. 

Correctness is crucial in the use of \ac{XAI} as explanations that do not align with the model's decision do not offer a meaningful way to assess the plausibility of an \ac{ML} decision. While explanations without Correctness may seem plausible, they can potentially increase automation bias \cite{Vered2023} or hinder proper scrutiny of the decision system \cite{Zhou2023}.
To evaluate a decision accurately, all factors influencing the decision should be highlighted. To be able to reject a decision based on sensitive attributes like gender, a combination of Correctness and Completeness is necessary.
Since most current \ac{XAI} methods provide no guarantee for Correctness (except for limited axiomatic properties in \ac{SHAP}), Consilience of the explanations of different methods could be used as a substitute.
It is important to note that even when multiple \ac{XAI} methods agree, Correctness of explanations cannot be guaranteed, just as human experts may make decisions based on reasons beyond their explanations.
In the case of human expert advice, the fiduciary assesses the plausibility based on their understanding of the expert's explanations.
Applied to \ac{ML}, this implies two criteria: explanations must be understandable to laypersons and to domain experts.
This also requires Consistency, as an inconsistent explanation method can create confusion for non-experts who cannot resolve differing explanations and hyperparameter choices.
Contrastivity is a necessary precondition to enable a user to discern between the prediction given by the model and relevant alternatives for action.
For Covariate Complexity, the same as for Correctness can be stated: Ideally, the explanations should consider features and feature interactions as complex as in the original model, while still being understandable for laypersons. But given that simplifications are made explicit and the person tasked with the plausibility check still considers an explanation plausible, these requirements can be loosened.

As described in Section~3.1, insufficient Compactness can render explanations unintelligible.
On the other hand, explanations should be as complete as possible to highlight all relevant features.
To resolve this conflict, each user should be shown the most complex explanation this user can still grasp, requiring an adaptation of the explanation to the Context.
One way to give an \ac{XAI} system this adaptability would be to make it controllable, allowing each user to adapt the explanation to their needs, e.g., by limiting the number of features highlighted in an explanation.
Controllability can resolve such conflicts and additionally increase the amount of time a user spends with an explanation, increasing the explanation’s effectiveness \cite{Cheng2019}.
Albeit Coherence to prior knowledge is not a requirement for explanations for plausibility checks, it will be a main part of the plausibility checks themselves, as laypersons and domain experts do not have other meaningful ways to assess \ac{ML} decisions and explanations than comparing them to their prior knowledge.

As the first requirement in the ISION rule states, the expert consulted must be an independent professional qualified for the issue to be clarified.
When extended to AI systems, the qualification of such systems for a general use case can be assessed via evaluation metrics such as accuracy, but the qualification for a specific data point with unknown label (as in real-world decisions) is more difficult to estimate.
One main possibility to assess whether a model is qualified to make a decision for a specific scenario is the quantification of the Confidence of the model.
Thus, a Confidence estimate of the model prediction should accompany the explanation for a specific decision.
As explanations are not guaranteed to be correct, a Confidence estimate of the explanation itself is necessary to assess whether it can be relied upon.

To facilitate the fiduciary's defense in potential lawsuits, explanations should be presented in a way that allows for permanent storage and thus, Constancy. In the case of automatically generated ambiguous explanations, e.g., saliency maps, fiduciaries are advised to document their interpretation to prevent later disputes.
Since the plausibility of a single decision needs to be assessed, local explanations should be sufficient.
To improve the understanding of the limits of an AI system, it could be helpful to “calibrate one’s trust” via continued interaction with a model, requiring more than one local explanation \cite{Ma2023}. 
Legally, \ac{XAI} could enable a comprehensive review of \ac{ML} decisions, surpassing the level of the ISION principles.
In legal discussions, there is debate about whether transparency requirements for algorithmic decisions should be no higher than those for human decisions.
However, equal treatment of humans and machines is not required by law. On the contrary: if it is technically feasible to provide better explanations for algorithmic decisions compared to human decisions, there is no reason not to utilize this greater efficiency and consequently take it as a legal prerequisite.

Summarizing, when extending the ISION-principles to non-human explanations, they do not limit the type of explanation that can be used to assess the plausibility of a decision.
While data attributions---showing similar decisions that happened in the past---could also be helpful here, it can be argued that such explanations are not specific enough to the decision at hand and require too much interpretation themselves.
Thus, the simplest way to aid in a plausibility assessment is to use explanations such as feature importance with axiomatic properties for Correctness and Completeness such as \ac{SHAP}, although the actual value of such properties is part of an ongoing discussion \cite{Kumar2020, Merrick2019, Sundararajan2018}.
As the requirement for Correctness is the most difficult one to fulfill, using multiple explanation methods (and thus, doing a “plausibility check” of the explanation itself) currently seems to be the best way to meet the legal requirements. 

\subsection{Perspective of Decision Recipients: A Right to Explanation} \label{ssec:right-to-explanation}

People are also increasingly becoming the recipients of algorithmic decisions, for example when \ac{ML} is used to decide whether someone receives a loan or is invited to a job interview.
As a result, decisions are being made on issues that are of central importance to social life and personal wealth.
Especially if the decision is negative, the person concerned has an interest in finding out why.
From a legal perspective, the question arises as to whether the person concerned has a ``right to explanation" of the decision and from which legal basis such a right can be derived. Of course, this raises the question of the extent of the legal guarantee, i.e., what must be disclosed/explained to the data subject.

It is being discussed whether the ``right to explanation” of algorithmic decisions can be derived from the provisions of the GDPR, specifically Article 15(1)(h) and Article 22(3) \cite{Wachter2016, Wachter2018, Malgieri.2017, Edwards2017, Selbst.2017, Dimitrova.2020, Kaminski.2018, Kaminski.2021, Kim.2018, Brkan.2019}. In our legal opinion, this is indeed correct.
Article 15(1)(h) grants the data subject the right to be informed of the existence of automated decision-making, including profiling, and to receive meaningful information about the logic, significance, and consequences of such processing.
This right is further supported by Article 22(3), which allows the data subject to contest AI decisions.
In the SCHUFA judgment, the Court of Justice of the European Union (CJEU) emphasized the need to interpret EU law not only of its wording but also of its context, objectives, and the purpose pursued by the act of which it forms a part \cite{CJEU2023}.
Given the GDPR's goal of enabling effective exercise of data subjects' rights, the “right to explanation” is already inherent in the GDPR. The Advocate General at the CJEU Priit Pikamäe, in his opinion on the SCHUFA case, shares a similar sentiment.
He considers that “the obligation to provide `meaningful information about the \emph{logic involved}' must be understood to include sufficiently detailed explanations of the method used to calculate the score and the reasons for a certain result. In general, the controller should provide the data subject with general information, notably on factors taken into account for the decision-making process and on their respective weight on an aggregate level \citep{GuideAutomatedDecision}, which is also useful for him or her to challenge any ‘decision’ within the meaning of Article 22(1) of the GDPR.” \cite{Pikamae2023}

Besides the GDPR, the right to explanation of algorithmic decisions may find grounding in fundamental rights.
At the EU level, considerations of human dignity and the right to the protection of personal data (Articles 1, 7 and 8 of the EU Charter of Fundamental Rights) come into play.
Article 8(2) sentence 2 of the EU Charter of Fundamental Rights affirms every person's right to access data collected about them and the right to have it rectified.
However, it is questionable whether this guarantee also includes the right to explanation of decisions that are the result of \ac{ML}-based data processing.
The narrow formulation of Article 8(2) sentence 2 may not explicitly encompass such rights.
Nevertheless, \ac{ML}-based data processing, like any data processing, constitutes an interference with the right to the protection of personal data, necessitating justification.
The proportionality test allows for weighing the data subject's interest in decision explanation against the \ac{ML} application operator's interest in protecting trade secrets (Article 16 and 17(2) of the EU Charter of Fundamental Rights, protection of freedom to conduct a business and of intellectual property).
The principle of practical concordance mandates finding a balance that is as gentle as possible, ensuring neither conflicting fundamental right takes precedence over the other (akin to Pareto optimum in game theory).
Thus, the goal is to explain the \ac{ML} prediction to the concerned individual without revealing algorithmic components constituting trade secrets.
However, the challenge of reverse engineering, often permitted in some national laws (e.g., in the German Law on the Protection of Trade Secrets), must be addressed, e.g., by new regulations. 

The task for \ac{XAI} regarding a “right to explanation” (independent of the legal basis used) can be summarized as: Enable end-users of AI systems to object to \ac{ML} decisions and---as argued by \citet{Wachter2017}---show ways to achieve a different decision.
These tasks require local explanations, as also argued by \citet{Edwards2017}.
There, they are called ``subject-centric" explanations, which include data-centric ones, which we do not recommend here because of privacy concerns when showing users other data instances.
As the arising requirements are similar to the ones in Section~\ref{ssec:plausibility-check}---due to the task of enabling a plausibility check of the decision by a layperson---only the differences are discussed here.

The GDPR grants users the right to object to a decision in Article 22(3) (here called Counterability), thus from a practical standpoint, a process for objecting automated decisions should be implemented.
This could also entail the option to object to explanations that are deemed unjust or incorrect, especially since Correctness cannot be guaranteed.
As much as a Confidence estimate of a decision and its explanation could help users to assess whether a model was certain in their case or whether it was more likely to be wrong, it could also complicate the privacy of such systems by enabling membership inference attacks \cite{Shokri2017}.
To prevent issues such as this, other measures could be taken, e.g., limiting the amount of explanation requests a user can make.
Similar problems arise for data influence explanations, as sensitive training data (other than their own) should not be shown to end-users.
Compared to fiduciary decisions, explanations can be generated when need arises, as it is not feasible (nor necessary) to compute explanations for all possible users in advance.

In our opinion, merely providing counterfactual explanations, as recommended in \cite{Wachter2017},  may not be sufficient to achieve the goal of enabling users to object to \ac{ML} decisions and give them the opportunity to alter the outcome.
This viewpoint is underscored by the statement of Advocate General Priit Pikamäe at the CJEU \cite{Pikamae2023}.
Because counterfactual explanations do only provide a very limited account of how a decision was made (“This feature must change to change the outcome, thus it is most likely important.”), further information about the decision needs to be given, which could for example take the form of feature importance explanations as in Section~\ref{ssec:plausibility-check}.
Thus, to fulfill all aims of the use of \ac{XAI} regarding the ``right to explanation'', several types of explanations could therefore be jointly given to users. 

\section{Legal Requirements: Model-Centric} \label{sec:model-centric}

Previously, the legal focus was on a single decision made by an \ac{ML} model.
In the next part, the focus is shifted to the model as a whole and the legal needs for \ac{XAI} in the areas of product safety and product liability are described, followed by a discussion of properties \ac{XAI} methods need to fulfill to truly aid in these cases. 

\subsection{Product Safety} \label{ssec:product-safety}
Under certain conditions, developers of \ac{ML} applications must comply with the requirements of the Product Safety Law.
The European Union’s regulations on product safety are outlined in the General Product Safety Directive (GPSD) of December 3, 2001, which is set to be superseded by the new General Product Safety Regulation (GPSR).
The GPSR came into effect on June 12, 2023, and its application will commence on December 13, 2024.
According to the definition in Art. 3(1) GPSR a “product” is any item, whether or not it is interconnected to other items, supplied or made available, whether for consideration or not, including in the context of providing a service, which is intended for consumers or is likely, under reasonably foreseeable conditions, to be used by consumers even if not intended for them.
It is still unclear whether “stand-alone” software should also be considered a product within the meaning of the GPSR.
According to the traditional view, a product requires a physical object.

Nevertheless, software becomes a product when installed on data carriers or embedded in a physical product.
In this case, at least the general requirement applies that only safe products can be placed or made available on the market (Art. 5 GPSR).
“Safe product” means any product which, under normal or reasonably foreseeable conditions of use, including the actual duration of use, does not present any risk or only the minimum risks compatible with the product’s use, considered acceptable and consistent with a high level of protection of the health and safety of consumers (Art. 3(2) GPSR).
This definition alone shows that absolute certainty is not owed; rather, the manufacturer must (only) take the measures that are objectively necessary and reasonable according to objective standards in the circumstances of the specific case.
Note that besides the manufacturer, another ``economic operator" would need to fulfill such duties. Such an ``economic operator" is defined in Art. 3(13) GPSR, e.g., as the authorised representative, the importer, the distributor, the fulfilment service provider, or any other natural or legal person who is subject to obligations in relation to the manufacture of products or making them available on the market in accordance with the GPSR.

Summarizing, the manufacturer or other ``economic operators" would have an increased interest in ensuring that their product, including the \ac{ML} application, is free of defects. 

\subsection{Product Liability (Tortious Liability)} \label{ssec:liability}
Product liability in the European Union is currently governed by the (almost 40-year-old) Directive 85/374/EEC on Liability for Defective Products.
However, on March 12, 2024, the European Parliament adopted the new EU Product Liability Directive, which is expected to enter into force soon.
This new directive treats software as a product (Art. 4(1) of the new directive) but does not apply to free and open-source software that is developed or supplied outside the course of a commercial activity.

The aim of product liability is to ensure that any natural person who suffers harm caused by a defective product has the right to claim compensation.
It is a tortious liability, not a contractual one.
The liability applies if the defective product causes death, personal injury, or damage to property other than the defective product, or destruction or corruption of data that is not used for professional purposes (Art. 5a(1) of the new directive).
This requires that the product is intended for private use or consumption and has been used accordingly.
In product liability law, liability does not normally apply if the defect could not have been recognized according to the state of the art in science and technology at the time when the product was placed on the market (so-called development risk).
However, the fact that the manufacturer of an \ac{ML} application cannot foresee its behavior in a specific hazardous situation does not constitute a development risk.
The blanket exoneration of manufacturers of AI systems is therefore out of the question.

The new directive also extends the requirements for freedom from defects: According to Article 6(1)f, a product is also defective if it does not fulfill the security-relevant cybersecurity requirements. At the same time, the new directive specifies the requirements for the exemption from liability.
The liability is excluded if the objective state of scientific and technical knowledge at the time when the product was placed on the market, put into service, or in the period in which the product was within the manufacturer’s control was not such that the defectiveness could be discovered (Art. 10(1)e of the new directive).
However, the manufacturer must ensure that both the software and products containing software are defect-free throughout the duration of their control.
During this period, the manufacturer must also provide the updates and upgrades necessary to maintain safety (Article 10(2) of the new directive).
“Control of the manufacturer” means the fact that the manufacturer of a product authorizes: a) the integration, connection, or supply of a component, including software updates or upgrades, by a third party or b) the modification of the product (Article 4(5) of the new directive).

It should be noted that there are unresolved legal issues with regard to software defects.
In the case of construction defects, for example, the question arises as to the standard of comparison: does an AI-steered vehicle already qualify as defective if it causes an accident that a human would have avoided?
Opponents argue that such an “anthropocentric concept of defect” does not do justice to the special features of digital products.
After all, autonomous vehicles are expected to reduce the number of accidents significantly. 
A separate “system-related concept of defect” must therefore be found for such products, which gives the manufacturers of autonomous systems effective incentives to improve the safety features of their products without demanding absolute safety and imposing the overall burden of accident costs on them \cite{geistfeld2017}.

In addition to the European product liability regime, there is also tortious product liability in Germany in accordance with Section~823(1) of the German Civil Code.
Here, the manufacturer is liable for damages to anyone whose legally protected interests (in particular life, health and property) have been injured by a defective product.
This presupposes that the manufacturer has breached marketing obligations (design, manufacturing, instruction, product monitoring, warning and recall obligations).
In the context of Section~823(1) of the German Civil Code, it is recognized that the manufacturer is obliged to constantly adapt the software to new knowledge about safety risks and to exploit new technical possibilities.

\subsection{Requirements for XAI} \label{ssec:safe-liable-XAI}
The above Sections~\ref{ssec:product-safety} and \ref{ssec:liability} result in the same task for \ac{XAI}:
Find and, if possible, rule out defects in the \ac{ML} model. \ac{XAI} methods are often stated to help in debugging models or showing the fairness of such systems \cite{Langer2021, Deck2023}.
Different solutions can be devised, ranging from global explainability of a model to rule out defects, to using local explanations or data attributions to find defects, to possibly not using \ac{XAI} methods at all given sufficient testing and documentation.
Ways in which \ac{XAI} can assist to find or rule out defects are covered in the following, for a way to ensure safety by documentation see \cite{Brajovic2023}.
In all of these cases, explanations can be expected to be generated for expert users.

To ensure freedom of defects, Correctness is at the heart of \ac{XAI} use, as a global explanation method that does not conform to the model might miss out on defects.
To date, no such global explanation method seems to be known and computationally feasible, as highlighted in the closely related field of formal verification of neural networks \cite{Albarghouthi2021}.
The only way to provide an explanation with such properties is to use white-box models, as demanded by \citet{Rudin2019} for safety critical use cases.
To spot defects within black-box \ac{ML} applications, local explanations can help, where Correctness-problems could once again be mitigated by the Consilience of multiple \ac{XAI} methods, although this would still not give providers of such system enough information to guarantee freedom of defects.
As \ac{XAI} in the near future is more likely to show the presence of defects than to provide correct global explanations on which a guarantee for the freedom of defects could be built, the first case will be considered for the following properties.

As long as the explanations do show relevant defects, there is no need for Complete explanations, as for instance even a very specific method, which only highlights whether a protected attribute is used in a decision, could help to spot unwanted behavior.
The same can be argued for Compactness, as not entire explanations need to be understood but they can be searched for parts where defects are suspected.
In practice, it could prove difficult to choose the (number of) features for such explanations.
In the search for defects, inconsistent explanation methods can be used, as executing them multiple times seems feasible to potentially spot defects.
Additionally, they do not necessarily need to provide Continuity, unless a global understanding of a model should be gained, which is only possible if explanations are continuous and can thus be generalized to near data points.
Since defects, for which liability could become relevant, e.g., discriminating decisions, can also result from the interaction of input features for instance via confounding variables, used \ac{XAI} methods should provide the maximum understandable amount of Covariate Complexity.
Additional to using \ac{XAI}, confounders could be discovered via methods as described by \citet{Rogozhnikov2022} to ensure that no sensitive attributes are reconstructed based on other input features \cite{Wachter2018}.
A Confidence estimate could itself show unwanted behavior in a model, e.g., a model being uncertain about a safety-critical decision that is expected to arise during its deployment.
Because of this, model Confidence should be considered.
Since the Controllability of an explanation improves the user’s understanding of the underlying model and thus, the chance of finding relevant defects, explanations should be interactive.
To avoid liability claims and prove a freedom of defects (or the safety of the AI system), the relevant explanation artifacts and the way in which they were created need to provide Constancy.

Overall, there is no clear consensus on how \ac{XAI} can help to spot or rule out defects in \ac{ML} applications.
While a human intelligible and fully correct global explanation seems ideal, it also seems unrealistic for some use cases as of now---partly due to the currently available \ac{ML} and \ac{XAI} methods, but also due to the complexity of some use cases.
The closest to this ideal are interpretable models, for which guarantees can be derived, given that they reason on human-intelligible features.
If no such model can be devised for a certain use case, the best way forward seems to be a joint combination of documentation and a variety of explanation methods for the model itself, its decisions and the data used.

\begin{table*}[htb]
\small
\centering
\begin{tabular}{|p{3.2cm}|p{3.2cm}|p{3.2cm}|p{3.2cm}|}
    \hline
    \textbf{Legal Basis} & \textbf{Fiduciary Decisions - Section~\ref{ssec:plausibility-check}} & \textbf{Right to Explanation - Section~\ref{ssec:right-to-explanation}} & \textbf{Safety/Liability - \newline Section~\ref{sec:model-centric}} \\
    \hline\hline
    \ac{XAI}-Task & Check the ML decision for plausibility & Enable user to understand and object to decision & Show the absence of errors (as far as possible) \\
    Audience of Explanation & Laypersons, domain experts & Laypersons & ML experts \\\hline
    Correctness & $\times(1)$& $\times(1)$ & $\times(1)$ \\
    Completeness & $\times(2)$ & $\times(2)$ &  \\
    Consistency & $\times$ & $\times$ &  \\
    Continuity & & & \\
    Contrastivity & $\times$ & $\times$ &  \\
    Covariate Complexity & $\times(3)$ & $\times(3)$ & $\times$ \\
    Compactness & $\times(2)$ & $\times(2)$ &  \\
    Compositionality & & & \\
    Confidence & $\times$ (for prediction and explanation) & $\times$ (for prediction and explanation) (4) & $\times$ (for prediction) \\
    Context & $\times$ & $\times$ &  \\
    Coherence & & & \\
    Controllability & $\times$ & $\times$ & $\times$ \\
    Consilience & $\times(1)$ & $\times(1)$ & $\times(1)$ \\
    Computations & & & \\
    Coverage & Prediction (5), Data & Prediction & Model, Prediction, Data \\
    Counterability &  & $\times$ &  \\
    Constancy & $\times$ &  & $\times$ \\\hline\hline
    Recommendation for \newline Explanation Type & Feature Importance (or similar) & Counterfactuals and Feature Importance (or similar) & Global explanations or white-box models \\
    \hline
\end{tabular}
\caption{Requirements for \ac{XAI} derived from the legal areas. $\times$ denotes that a property is necessary for that specific use case. \newline (1) denotes that instead of Correctness, Consilience of explanations can be used. \newline (2) denotes a tradeoff between Completeness and Compactness, \newline (3) denotes a tradeoff between Completeness/Covariate Complexity and understandability, \newline (4) denotes a tradeoff between Confidence estimates and privacy. \newline (5) denotes that multiple predictions could be used to ``calibrate one's trust", e.g., to learn when to trust the ML model.}
\label{tab:summary}
\end{table*}

\section{Discussion} \label{sec:discussion}
While the EU AI Act is seen as a breakthrough in AI regulation, it is important to note that it is not expected to supersede all other relevant laws influencing the use of AI (and thus, \ac{XAI}).
The laws described in Sections~\ref{sec:decision-centric}~and~\ref{sec:model-centric}, from which the requirements were derived, mainly pertain to civil liability and will coexist with the AI Act, which falls under public law.
Although these laws interact (unfulfilled requirements of the AI Act might be seen as defects in the sense of Section~\ref{ssec:liability}), any further discussion on their possible interactions is beyond the scope of this paper.

There has been much debate about the perceived conflict between regulation and innovation in the context of the AI Act.
However, in our opinion, regulation itself does not stifle innovation; rather, unclear regulation or instructions may impede progress.
To avoid such ambiguity in the field of \ac{XAI}, this paper proposes a structured approach based on specific properties to guide \ac{XAI} use and development.
Similarly, international or national norms could recommend \ac{XAI} properties or methods for specific use cases, providing clear instructions on which methods to employ under different circumstances.
These norms should consider additional legal foundations for \ac{XAI}, such as criminal, administrative, and insurance law.

To promote advancements in the intersection of law and \ac{XAI}, technical and legal experts should actively exchange information.
While legal experts often emphasize the need for transparency to ensure safe AI usage and address liability issues, the current technical possibilities do not fully support such a broad statement.
Thus, the law must be informed about the strengths and weaknesses of different \ac{XAI} methods in order to be able to adapt its requirements if necessary. Since technological development does not halt, a continuous dialogue between law and XAI research is required.
This dialogue could also address communication challenges resulting from the different terminology used in both fields \cite{Schneeberger2023}.
While the collaboration and integration of both fields are important, it is worth noting that the law cannot fully prescribe the use of technology.
In the case of \ac{XAI}, other disciplines such as social sciences (examining the effectiveness of explanations) and user interface design (determining how explanations should be presented for desired effects) can contribute to filling gaps in legal requirements.
As the law only provides minimum requirements, insights from these fields become more crucial for properties not touched by legal requirements (such as Compositionality and others) and when surpassing these minimum standards.

\section{Summary} \label{sec:summary}

In this paper, the relationship between law and \ac{XAI} was explored, deriving requirements for different areas of law based on an extended taxonomy of \ac{XAI} properties.
The areas of law considered were fiduciary decisions, GDPR, and product safety and liability.
These areas require different \ac{XAI} properties and methods.
For fiduciary decisions, assessing plausibility is important, similar to the GDPR, where users should be able to object to decisions and receive actionable advice.
In product safety, identifying defects in \ac{ML} models is crucial, where showing the absence of defects would be ideal.
A tabular summary of the derived requirements and explanation recommendations can be found in Table~\ref{tab:summary}, serving as a guidance for working with \ac{XAI} in these areas.

Various potential aims for \ac{XAI} have been proposed in the literature (see for example \citet{Saeed2023}), but the practical utilization of \ac{XAI} to achieve these goals remains unclear.
The lack of consensus regarding evaluation metrics for the objectives and properties of \ac{XAI} further complicates the matter, as the reliable and valid assessment of \ac{XAI} method characteristics is a necessary precondition to meaningfully confirm or reject specific methods for a given use case.
Possible actions to address this could involve developing new evaluation metrics or conducting studies to assess the effectiveness of \ac{XAI} in specific settings.
However, while the legal practice often requires the use of state of the art technologies, it appears that the current state of \ac{XAI} is not fully equipped to fulfill all legal requirements.

The most challenging \ac{XAI} properties in this regard are Correctness (explanations conforming to model logic), Confidence estimates of models and explanations themselves, and the lack of global explanation methods, indicating the need for further research in these areas.
An additional study by some of the authors focused on the use of \ac{XAI} for safe AI development and AI certification \cite{fresz2024XAIinCertification}.
It showed that practitioners in these areas view XAI as a practical asset to spot errors in ML models but do not expect XAI methods to provide comprehensive solutions to AI certification, due to similar problems to the ones mentioned above.

To inform future development, law can be used as an inspiration for \ac{XAI} methods with specific properties or new ways to evaluate \ac{XAI}.
To this end, we urge practitioners from both fields to collaborate, so that a common terminology and a common understanding of the capabilities of \ac{XAI} can be established.
The requirements for \ac{XAI} presented in this paper serve as recommendations that contribute to the ongoing discussion on the utilization and future development of \ac{XAI}.
We eagerly anticipate the future advancements in this field, both from a technical standpoint and within the legal and normative realms.

\section*{Acknowledgments}
Parts of this paper were created with the help of a company-specific implementation of ChatGPT (3.5 turbo).
It was used to create LaTeX-code for tables and to refine the drafts of some sections.
We thank all reviewers of the previous versions of this paper, who helped to improve it with their valuable feedback.

This paper is funded in parts by the German Federal Ministry for Economic Affairs and Climate Action under grant no. 19A21040B (project ``veoPipe'') and by the Fraunhofer Gesellschaft under grant no. PREPARE 40-02702 (project ``ML4Safety").

\bibliography{aaai24}

\end{document}